\documentclass[letterpaper]{article} 
\usepackage[preprint]{aaai2027} 
\usepackage[hyphens]{url}  
\usepackage{graphicx} 

\usepackage{listings}
\usepackage{xcolor}

\usepackage{natbib}  
\usepackage{caption} 
\usepackage{algorithm}
\usepackage{algorithmic}
\usepackage{amsmath} 
\usepackage{amssymb}
\usepackage{algorithm}
\usepackage{algorithmic}
\usepackage{booktabs}
\newcommand{\spm}[1]{\textsubscript{$\pm$#1}}
\usepackage{multirow}
\usepackage{pifont}
\usepackage{graphicx}
\usepackage{glossaries}
\usepackage{bm}
\usepackage{newfloat}
\usepackage{listings}
\DeclareCaptionStyle{ruled}{labelfont=normalfont,labelsep=colon,strut=off} 
\floatstyle{ruled}
\newfloat{listing}{tb}{lst}{}
\floatname{listing}{Listing}

\usepackage{booktabs}

\title{Neuro-Symbolic Predicate Learning for Semantic Safe Robot Control}
\author{
    Zihan Ye \textsuperscript{\rm 1,2,5}\equalcontrib \ , Jiayi Liu\textsuperscript{\rm 1}\equalcontrib,\ Puze Liu\textsuperscript{\rm 7}\corresponding, Jiayun Li\textsuperscript{\rm 1,4,5}, \\ Georgia Chalvatzaki\textsuperscript{\rm 1,4,5,8},\ Jan Peters\textsuperscript{\rm 1,3,5,6,8},\ Kristian Kersting\textsuperscript{\rm 1,2,5,6} 
}
\affiliations{
    \textsuperscript{\rm 1}TU Darmstadt, \textsuperscript{\rm 2}AIML group, \textsuperscript{\rm 3}IAS group,
    \textsuperscript{\rm 4}PEARL group,\\
    \textsuperscript{\rm 5}Hessian AI,
    \textsuperscript{\rm 6}DFKI,
    \textsuperscript{\rm 7}Tongji University
    \textsuperscript{\rm 8}Robotics Institute Germany\\

    zihan.ye@tu-darmstadt.de, puze\_liu@tongji.edu.cn
}

\newcommand{\cmark}{\ding{51}}
\newcommand{\xmark}{\ding{55}}
\newcommand{\pmark}{\ding{108}}

\definecolor{softred}{HTML}{F7623B}
\definecolor{softblue}{HTML}{458FB4}
\definecolor{softgreen}{HTML}{008000}
\newcommand{\prolog}{\mathrel{:\!\!-}}

\newacronym{vlm}{VLM}{Vision-Language Model}
\newacronym{vla}{VLA}{Vision-Language-Action Model}
\newacronym{fol}{FOL}{First-Order Logic}
\newacronym{neupro}{NEUPRO}{\underline{Neu}ro-Symbolic \underline{P}redicate Learning for Semantic Safe \underline{Ro}bot Control}
\newacronym{lpm}{LPM}{Logical Predicate Model}

\begin{document}

\maketitle

\begin{abstract}
As robots are increasingly deployed in everyday environments, ensuring their safety has become a central challenge. Existing methods often encode safety requirements as opaque mathematical/logical formulations or dense cost functions. While effective in specific tasks, they remain difficult to interpret, tightly coupled to individual tasks, and offer limited insight into why a robot action is considered safe or unsafe. To address this limitation, we propose ``Neuro-Symbolic Predicate Learning for Semantic Safe Robot Control'' (NEUPRO), which leverages a differentiable reasoner that can learn reusable safety representations from human-specified safety knowledge. NEUPRO allows practitioners to express task-related safety requirements as transparent symbolic rules, while enabling gradients to propagate through these rules to a feature extractor that maps raw observations to safety-relevant concepts. As a result, the learned feature extractor is (softly) grounded in human-understandable semantics, supports transparent constraint evaluation, and is transferable across tasks. By coupling interpretability with differentiability, NEUPRO moves beyond opaque cost design toward reusable safety reasoning. To evaluate NEUPRO's capability, we collect and release REASON, the first real robot benchmark dataset for interpretable robot safety specification. Experiments on REASON show that NEUPRO learns safety-critical features that generalize across tasks, mitigate the interpretability limitations of conventional black-box cost formulations, and provide explicit explanations of safety violation.
\end{abstract}


\begin{figure}
    \includegraphics[width=\columnwidth]{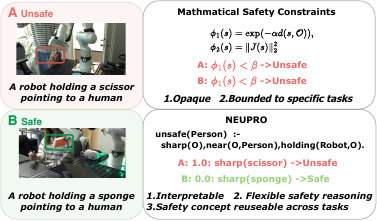}
    \caption{Existing safe robot learning methods often encode safety constraints using opaque mathematical/logical functions, which are hard to interpret and bound to specific tasks. We propose NEUPRO, leveraging differentiable reasoning to alleviate these limitations using interpretable symbolic rules. 
    }
    \label{fig:teaser}
\end{figure}

\section{Introduction}
As robots are increasingly expected to operate in complex, unstructured environments, ensuring safety has shifted from a desirable attribute to a fundamental requirement~\citep{garcia2015comprehensive,achiam2017constrained,wachi2024survey,brunke2022safe,zhao2023state}. Whether navigating around humans~\citep{lasota2017survey} or manipulating fragile objects, a robot needs to continuously reason about underlying safety constraints~\citep{haddadin2017robot}. Safe robot learning aims to address this by enabling policies to acquire optimal behaviors while strictly respecting these boundaries. However, a central challenge remains: how should safety requirements be constructed and transferred across diverse tasks? 

Existing safe robot learning methods typically encode safety requirements via predefined logical rules or mathematical constraints~\citep{brunke2025semantically,liu2023safe,liu2025safe}. While effective, these representations face several important limitations. 
Hand-crafted logical rules and mathematical constraints can provide strong safety guarantees, but they are typically designed for specific tasks and require substantial engineering effort to formulate and adapt. This difficulty becomes more pronounced for abstract, visually grounded requirements, such as ``do not point a knife at a human,'' which cannot be easily translated from raw images into explicit mathematical constraints. 
Learning-based approaches~\citep{kim2023learning,lindner2024learning, gunster2024handling} may reduce the need for manual specification, but the resulting safety representations are often entangled with the task and policy, lacking interpretability and limiting their potential reuse in new settings.
\gls{vlm}s offer a promising alternative because they can interpret natural-language safety instructions directly from visual observations. However, their predictions are generally difficult to verify and do not, by themselves, provide formal safety guarantees. 
Recent work has begun to incorporate semantic scene understanding into robot safety mechanisms~\citep{brunke2025semantically}, this semantic information is still translated into task-specific mathematical constraints. The central challenge is therefore to ground human-interpretable safety requirements directly from low-level visual observations while preserving sufficient structure for verification and integration into safe robot learning. 


Driven by the need that safety constraints should not only be optimized but also understood, we propose \gls{neupro}. \gls{neupro} allows one to construct task-relevant safety constraints as transparent symbolic rules with flexible compositions of learnable logical predicates. 
To train the predicates, a differentiable logic reasoner enables gradients to flow from the safety-rule satisfaction signal back to \gls{lpm} inferred from a visual foundation model, such as Grounding DINO~\cite{liu2023grounding}. Consequently, the \gls{lpm} learns to map raw observations into the grounded atoms required by the symbolic rules. Through this formulation, the learned representation is not merely optimized to fit an opaque label, but is explicitly (softly) grounded in interpretable safety concepts. Furthermore, such \gls{lpm} allows us to achieve object-level and task-level generalization without retraining. In our setting, we assume the safety rules are predefined by domain expert as the normative decision must always come from the responsible institution.

To evaluate \gls{neupro}'s capability, we collect the REASON benchmark. To the best of the author's knowledge, REASON is the first real robot interpretable safety dataset. 
It covers safety scenarios involving human–robot interactions and robot–environment interactions, providing structured annotations that connect perceptual observations with interpretable safety constraints. The results show that \gls{neupro} can effectively learn interpretable safety-relevant predicates. The learned features can generalize to novel tasks that share safety semantics. Moreover, in contrast to conventional black-box cost formulations, \gls{neupro} can provide explicit explanations for safety violation. 

To summarize, our main contributions are:
\begin{itemize}
    \item We propose \emph{\gls{neupro}}, a neuro-symbolic framework that represents safety specifications as interpretable symbolic rules, facilitating flexible, interpretable safety reasoning directly from visual inputs.
    \item \emph{\gls{neupro}} supports scalable semantic-safety predicate learning and reasoning through its graph-based inference.
    \item We release \emph{REASON}, to the best of our knowledge, the first real robot interpretable safety specification dataset.
    \item We validate \emph{\gls{neupro}}'s effectiveness in safety predicates learning and safety reasoning on REASON, and demonstrate its interpretability, flexible reasoning, and safety predicate cross-task generalization capability over conventional safety constraints formulations.
\end{itemize}

\section{Related Work}
\label{sec:relatedwork}
\gls{neupro} builds on differentiable logic and is closely related to safe and neuro-symbolic robot learning. 

\textbf{First-order logic (FOL) and Differentiable Forward-Chaining Reasoning.} We refer readers to App.~A for a review of first-order logic fundamentals. Building on this foundation, differentiable forward-chaining inference~\citep{Evans18,Shindo21aaai,ye2022differentiable} enables logical entailment to be computed in a differentiable manner via tensor-based operations. This paradigm bridges symbolic reasoning with gradient-based learning, allowing logic-driven models to be trained end-to-end. However, despite their differentiability, tensor-based reasoning operations are inherently memory-intensive. To address this bottleneck, recent work such as \textit{Neumann}~\citep{Neumann} leverages Graph Neural Networks (GNNs) to mitigate memory overhead. While \gls{neupro} similarly utilizes a graph architecture for memory efficiency, our approach differs in focus. Whereas \textit{Neumann} focuses on inducing the logical rules, \gls{neupro} leverages graph-based reasoning to learn the neural feature extractor, focusing instead on \textit{representation learning} to softly ground high level concepts in raw perception. 

\begin{table}[t]
\centering
\label{tab:method_comparison}
\begin{tabular}{lccc}
\toprule
\textbf{Method} 
& \textbf{Interp.} 
& \textbf{Veri.} 
& \textbf{Flexi.}  \\
\midrule
Math. constraints 
& \textcolor{softred}{\xmark}
& \textcolor{softgreen}{\cmark} 
& \textcolor{softred}{\xmark} \\

Learning-based 
& \textcolor{softred}{\xmark} 
& \textcolor{softred}{\xmark} 
& \textcolor{softblue}{\pmark}  \\

VLM-based 
& \textcolor{softblue}{\pmark}
& \textcolor{softred}{\xmark} 
& \textcolor{softgreen}{\cmark} \\

\textbf{\gls{neupro}} 
& \textcolor{softgreen}{\cmark} 
& \textcolor{softgreen}{\cmark} 
& \textcolor{softgreen}{\cmark} \\
\bottomrule
\end{tabular}%
\caption{NEUPRO alleviates the limitations of existing safety reasoning paradigms by supporting interpretable, verifiable, and flexible safety reasoning. We compare different paradigms along these three dimensions, where \textcolor{softgreen}{\cmark}, \textcolor{softblue}{\pmark}, and \textcolor{softred}{\xmark} indicate strong, partial, and limited support, respectively.}
\end{table}

\textbf{Neuro-Symbolic Robot Learning.}
A central line of neuro-symbolic robot learning uses symbolic predicates and operators to bridge continuous robot observations with high-level task and motion planning. For example,~\citet{silver2022} learn neuro-symbolic skills that integrate symbolic operators with neural policies for bi-level task and motion planning. ~\citet{chitnis2022learning} learn neuro-symbolic relational transition models for planning in continuous robotic domains, while VisualPredicator~\citep{liang2025visualpredicator} learns task-relevant predicates and abstract world models from robot interaction data to improve planning and generalization. Other approaches, such as Dylan~\citep{ye2025learning} and NeSy Plan~\citep{keller2025neuro}, use symbolic abstractions to decompose long-horizon tasks into reusable skills and high-level plans. Unlike these methods primarily use symbolic structures for task planning and skill composition,  \gls{neupro} uses symbolic rules to reason \emph{whether a scene satisfies safety requirements and why}. Specifically, \gls{neupro} represents safety requirements as interpretable symbolic rules and uses a differentiable reasoner to ground safety-relevant predicates from raw visual observations. This allows safety supervision to shape the learned perceptual representation while preserving explicit explanations of constraint satisfaction.

\begin{figure*}
\centering 
    \includegraphics[width=\textwidth]{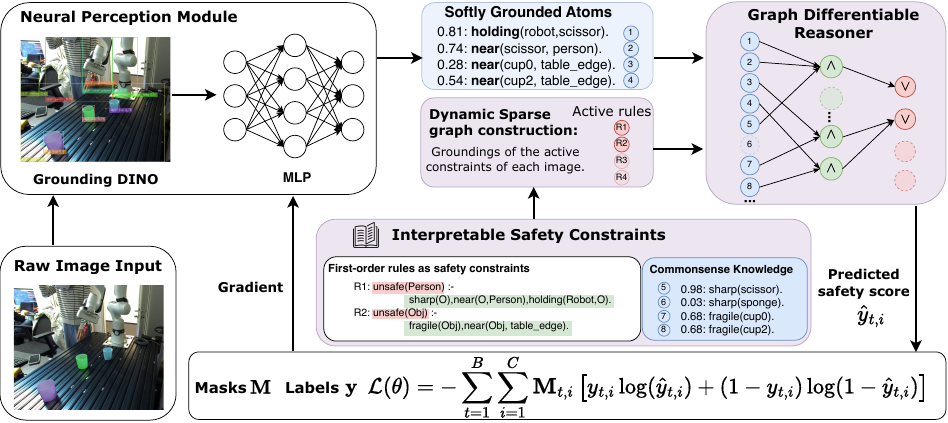}
\caption{\textbf{Overview of \gls{neupro}.} \gls{neupro} supports interpretable safety reasoning, end-to-end grounding of safety concepts in images, and flexible reasoning with commonsense knowledge. \gls{neupro} consists of two key components: (i) a neural perception module and (ii) a graph-based differentiable reasoner. Given a raw image, the perception module uses Grounding DINO and a learnable MLP to extract object-centric features and predict soft truth values for safety-relevant grounded atoms. The differentiable reasoner then infers the safety prediction based on the ground atom evaluation, safety rules, and commonsense knowledge. 
For details, please see Sec.~\ref{sec:method}.  
}
    \label{fig:framework}
\end{figure*}

\textbf{Safe Robot Learning.} 
A common approach in safe robot learning is to specify safety constraints manually. For example, in constrained reinforcement learning~\citep{achiam2017constrained,liu2022constrained,yu2022reachability}, safety constraints are encoded as a cumulative cost, and task performance is optimized subject to a safety budget. Other methods enforce safety through control-theoretic mechanisms, including control barrier functions~\citep{ames2019control}, reachability analysis~\citep{selim2022safe,ganai2023iterative}, and shielding~\citep{yang2023safe,alshiekh2018safe}. Recent methods further explore differentiable barrier-function architectures~\citep{xiao2023barriernet,xiao2025abnet} or embed safety constraints into constraint manifolds~\citep{liu2022robot,liu2023safe,liu2025safe}. Their safety constraints are typically encoded as opaque scalar costs or mathematical representations. Such formulations are often difficult to interpret semantically and are usually tied to specific tasks. 
Learning-based approaches~\citep{kim2023learning,lindner2024learning} may reduce the need for manual specification, but the resulting safety representations are often entangled with the task and policy, lacking interpretability and limiting their potential reuse in new settings. In contrast, \gls{neupro} expresses safety specifications as interpretable FOL rules and supports flexible safety reasoning and safety feature reuse.
\section{Neuro-Symbolic Predicate Learning for Semantic Safe Robot Control (NEUPRO)}
\label{sec:method}
Let us now introduce \gls{neupro}, a framework for learning safety-aware perceptual representations from raw observations while preserving the structure and interpretability of symbolic safety rules. As summarized in Fig.~\ref{fig:framework}, \gls{neupro} consists of two main components. \textbf{(i)} A \glsentryfull{lpm} maps raw images to soft truth values of grounded atoms. \textbf{(ii)} A graph-based differentiable reasoner evaluates first-order safety rules through index-based message passing over dynamically instantiated rule groundings. Together, they enable memory-efficient end-to-end learning, avoiding dense tensor-based differentiable reasoning.

\subsection{Problem Formulation}
The overall goal is to represents safety requirements as a set of interpretable \gls{fol} rules
$\{r_1, r_2, \dots, r_C\}$. Each rule $$r_i \prolog p_m(t_k, \dots), \dots, p_n(t_l, \dots).$$ is defined over a predicate vocabulary $p_m, p_n \in \mathcal{P}$ and a set of terms $t_k, t_l \in \mathcal{T}$, including object variables and constants. Specifically, 
A \textit{Language} $\mathcal{L}$ is a tuple of $(\mathcal{P}, \mathcal{A}, \mathcal{V})$, where $\mathcal{P}$ is a set of \textit{predicates}, $\mathcal{A}$ is a set of \textit{constants}, 
and $\mathcal{V}$ is a set of variables. A \textit{term} is a constant or a variable
. A \textit{ground term} or simply a \textit{fact} is a term with no variables. We denote an $n$-ary predicate $p$ by $p/n$. An \textit{atom} is a formula $p(t_1, \dots, t_n)$, where $t_1, \dots, t_n$ are terms. A \textit{literal} is an atom or its negation. A \textit{clause} is a finite disjunction $\lor$ of literals. A \textit{definite clause} is a clause with exactly one positive literal, such as $A\lor \lnot B_1 \lor \dots \lnot B_2$. We can write definite clasues in the form of $A\prolog B_1, \dots, B_n$.


NEUPRO assumes that the the symbolic safety rules are given. That is, our goal is not to learn the rules themselves, but to learn how to ground their predicates $p_i$ from raw robot observations. However, training each predicate classifier independently requires explicit labels for every predicate in every scene, which becomes costly as the number of objects, relations, and rules grows. Instead, we uses differentiable rule evaluation as a structured supervision signal: even when only rule-level labels or partially annotated predicates are available, gradients can propagate through the symbolic rules to guide the \gls{lpm}. As a result, the learned predicates remain interpretable and reusable for the downstream safety-reasoning tasks. We can formulate the problem as:

Let $\mathcal{D} = \{(s_t, \bm{y}_t, \bm{m}_t)\}_{t=1}^{N}$
denote a dataset of scene observations, where $s_t \in \mathcal{S}$ is a raw observation, such as an image. The vector $\bm{y}_t \in \{0,1\}^{C}$ contains rule-level safety labels, where $y_{t,i}=1$ indicates that observation $s_t$ satisfies the $i$-th safety rule and $y_{t,i}=0$ indicates a violation. The mask $\bm{m}_t \in \{0,1\}^{C}$ specifies which safety rules are active.
Given an observation $s_t$, our goal is to learn a neural \gls{lpm}
\[
    p_{\theta, j}(s_t) = P(g_j = \mathrm{True} \mid s_t; \theta)
\]
where $g_j \in \mathcal{G}$ denotes the grounded atoms, such that the induced rule evaluated by a differentiable logical reasoner $\mathcal{R}(\cdot)$ support accurate and interpretable prediction of rule satisfaction, i.e., $\mathcal{R}(p_\theta(s_t)) \odot \bm{m}_t \simeq \bm{y}_t \odot \bm{m}_t$.

\subsection{Perceptual Symbol Grounding}

\gls{neupro} leverages Grounding DINO~\cite{liu2023grounding} to extracts a set of object features from raw image batch. Base on the labeled rules associated with the image, we select a subset of features $\bm{s} \in \mathbb{R}^{|\mathcal{G}| \times N_f}$ whose object's class matches the terms appearing in the rule. The features are flattened over batch, $\mathcal{G}$ denotes the set of all grounded atoms in the batch, and $N_f$ is the feature dimension. The robot then infers the truth values of safety-critical predicates, such as whether a collision is likely. To this end, a neural \gls{lpm} $p_\theta$ outputs a continuous valuation vector
\begin{equation}
    \bm{v} = p_\theta(s) \in (0,1)^{|\mathcal{G}|}.
\end{equation}
Each entry $v_{i, j}$ corresponds to the soft truth value of a grounded atom in data point $i$.
For example, a grounded atom \(g_j\) may take the form
\[
    0.9: \mathtt{collision}(\mathtt{robot}, \mathtt{obstacle}).
\]
indicating that the \(\mathtt{robot}\) has a \(0.9\) probability of colliding with the \(\mathtt{obstacle}\). These continuous valuations serve as the initial valuation states for the differentiable reasoning.

\subsection{Reasoning Graph for \gls{neupro}} 

To alleviate the memory bottleneck of tensor-based differentiable reasoning, \gls{neupro} dynamically constructs a sparse reasoning graph for each mini-batch of the rule groundings of the active safety specifications. The graph contains three types of node, a \textit{Grounded Atom Node} computes the soft truth value for the corresponding predicates for each instance (e.g., $\mathtt{0.81:holding(robot,scissor}$); a \textit{Conjunction Node} computes the logical conjunction probability among multiple grounded atom (e.g., 
$\mathtt{fragile(cup0)} \land \mathtt{near(cup0, table\_edge)}$); and a \textit{Disjunction Node} computes logical disjunction probability among multiple applied instances, such as $\mathtt{unsafe(cup0)} \lor \mathtt{unsafe(cup2)}$. An illustrative figure can be found in the Differentiable Reasoner Graph block in Fig.~\ref{fig:framework}. 

\textbf{Literal polarity} is handled through an affine transformation of the corresponding atom valuation, using sign $\bm{s}$ and bias $\bm{b}$. For a positive literal, \gls{neupro} uses $[\mathrm{sign}, \mathrm{bias}] = [1,0]$, leaving the valuation unchanged. For a negated literal, it uses $[\mathrm{sign}, \mathrm{bias}] = [-1,1]$, which maps a valuation $x$ to $1-x$. Using these structural tensors, the soft truth values of all grounded literals are computed in parallel as
\begin{equation}
    \bm{l} = \bm{v} \odot \bm{s} +\bm{b},
\end{equation}
where $\bm{l} \in \mathbb{R}^{|\mathcal{G}|}$ contains the valuation of the grounded literal.

\textbf{Tensor Indexing.} For every grounded literal, the graph structure of \gls{neupro} is computed through two indexing tensors: 
$\bm{i}_{c}\in \mathbb{N}^{|\mathcal{G}|}$ specifies the edge connecting the grounded literal $\bm{l}$ to the valid conjunction node using $\mathtt{logical\_and}$ operation $\land$, and 
$\bm{i}_{d}\in \mathbb{N}^{|\mathcal{G}|}$ specifies the edge connecting multiple conjunction node applied by the same rule to the disjunction node via $\mathtt{logical\_or}$ operation $\lor$. 

\textbf{Differentiable Aggregation.} Given the grounded literal valuations, \gls{neupro} performs two steps over the sparse reasoning graph:
\begin{description}
    \item[ 
\textbf{Step 1: From grounded literal to conjunction nodes.}]
All literals belonging to the same grounded clause are aggregated by a differentiable $\land$ using product t-norm:
\begin{equation}
    \mathbf{h}_{\mathrm{conj}}= 
    \mathrm{scatter\_prod}
    \left(\bm{1},
        \mathbf{l},
        \mathbf{i}_{\mathrm{c}},
        \mathrm{dim\_size}=N_{\mathrm{c}}
    \right),
\end{equation}
where $\bm{1}$ is a $N_\mathrm{c}$ dimensional ones vector and $N_\mathrm{c}$ is the total number of valid grounded conjunction nodes in the batch. The resulting value $\mathbf{h}_{\mathrm{conj}}$ represents the soft truth value of the corresponding grounded rule body.

\item[\textbf{Step 2: From conjunctions to rule outputs.}] Multiple grounded conjunctions may satisfy the same safety rule. Therefore, \gls{neupro} aggregates all clauses associated with the same rule head into a rule-level satisfaction probability. We apply a differentiable Soft-OR aggregation to the disjunction node as: $\bm{H}_{\mathrm{out}}=$
\begin{equation}
    \mathrm{scatter\_softor_\gamma }
    \left(
        \mathbf{h}_{\mathrm{conj}},
        \mathbf{i}_{\mathrm{d}},
        \mathrm{dim\_size}=B \times C
    \right).
\end{equation}
where $C$ is the total number of clauses in the batch and   $\mathrm{softor}^{\gamma}$ is a smooth logical {\em or}  function: 
\begin{align}
    \mathrm{softor}_\gamma(x_1, \ldots, x_n) = \gamma \log\sum\nolimits_{1\leq i \leq n} \exp(x_i / \gamma),
    \label{eq:softor}
\end{align}
where $\gamma > 0$ is a smooth parameter. Eq.~\ref{eq:softor} is an approximation of the \emph{max} function over probabilistic values based on the \emph{log-sum-exp} approach~\citep{cuturi2017soft}. The resulting vector $\mathbf{H}_{\mathrm{out}} \in [0,1]^{B \cdot C}$ stores the satisfaction probabilities for all batch-rule pairs.
\end{description}
Finally, we reshape $\bm{H}_{\mathrm{out}}$ into a rule-satisfaction matrix
\begin{equation}
    \mathbf{P} =
    \mathrm{reshape}
    \left(\bm{H}_{\mathrm{out}}, [B,C] \right),
\end{equation}
where each entry
 $\mathbf{P}[t,i]
    =
    \bm{H}_{\mathrm{out}}[t \cdot C + i]$
represents \gls{neupro}'s predicted probability that observation $s_t$ satisfies safety rule $F_i$.

Because the reasoning procedure is implemented using differentiable indexing and scatter operations, gradients can propagate from rule-level supervision back to the underlying perceptual grounding network. At the same time, the intermediate conjunction nodes and sparse edges preserve an interpretable correspondence between grounded predicates, rule bodies, and final safety predictions.

\subsection{Learning Objective}

\gls{neupro} is trained to match rule-level safety satisfaction. For each observation $s_t$ and safety rule $F_i \in \mathcal{K}$, let
$y_{t,i} \in \{0,1\}$
where $y_{t,i}=1$ indicates that $s_t$ satisfies rule $F_i$, and $y_{t,i}=0$ indicates a violation. \gls{neupro} predicts
\begin{equation}
    \hat{y}_{t,i}
    =
    P(F_i = \mathrm{True} \mid s_t; \theta)
    =
    \mathbf{P}[t,i].
\end{equation}

Since not every rule is active for every observation, we use an active constraint mask $ \mathbf{M} \in \{0,1\}^{B \times C},$
where $\mathbf{M}_{t,i}=1$ indicates that rule $F_i$ should contribute to the loss for observation $s_t$. The masked binary cross-entropy loss is then $\mathcal{L}(\theta) =$
\begin{equation}
\small
\begin{aligned}
    - \sum_{t=1}^{B} \sum_{i=1}^{C}
    \mathbf{M}_{t,i}
    \Big[
         y_{t,i} \log \hat{y}_{t,i} 
        + (1-y_{t,i}) \log (1-\hat{y}_{t,i})
    \Big].
\end{aligned}
\end{equation}
This objective encourages \gls{neupro} to assign high satisfaction probabilities to valid safety rules and low satisfaction probabilities to violated rules. As a result, the perception module learns representations that are directly shaped by symbolic safety semantics, while the differentiable reasoner provides an interpretable and end-to-end trainable bridge between observations and logical constraint satisfaction.

\section{Experiments}
\label{sec:result}
With \gls{neupro} at hand, we now evaluate \gls{neupro}'s capability. 
Specifically, we aim to answer the following research questions: 
\textbf{RQ1:} Can \gls{neupro} effectively learn safety-related predicates from images through reasoner supervision? \textbf{RQ2:} Can \gls{neupro} accurately identify safety violation from visual observations?
\textbf{RQ3:} Can \gls{neupro} provide explanations for safety violation? \textbf{RQ4:} Can learned predicates transfer to different objects and task configurations? \textbf{RQ5:} Can \gls{neupro} flexibly reason about safety violations, rather than relying on fixed object-relation associations? \textbf{RQ6:} Does \gls{neupro} enable scalable predicate learning and inference? 

\paragraph{Benchmark datasets.} 


Existing robot safety datasets, such as OopsieVerse~\citep {balaji2026oopsieverse}, often supervise models with trajectory-level risk scores or binary safe/unsafe labels. While useful for assessing whether a safety violation occurs, such supervision typically could not be reason which safety conditions is violated from visual input. This limits the evaluation of models that aim to ground and reason over human-understandable safety concepts. To address this gap, we collect and introduce \textbf{REASON}, a rule-grounded real-robot benchmark dataset for interpretable safety reasoning from visual observations. REASON covers robot--environment and robot--human safety scenarios. Each observation is paired with safety-relevant predicates, human-specified symbolic constraints, and safety labels, enabling evaluation of the full perception-to-reasoning pipeline: grounding safety-critical concepts from raw images, composing them through executable rules, and identifying the conditions responsible for each safety decision. To facilitate reproducible research, we will open-source REASON together with the annotation tool used to create its predicate annotations, symbolic constraints, and rule-level safety labels.
Collectively, these tasks require recognizing object properties, spatial configurations, manipulation states, and human-directed object motion. See App.~B for a detailed description of individual tasks. 

\paragraph{Experimental Setup and evaluation metrics.} For each task, \gls{neupro} receives a RGB image observation and predicts safety scores. The safety score is used both as a safety prediction and as a learning signal for the visual feature extractor. We evaluate safety recognition using accuracy, precision, recall and F1 score as our evaluation metric. 

\paragraph{Baselines.} We compare \gls{neupro} with six baselines, including: 
\textbf{a Black-box classifier:} a neural classifier trained end-to-end to predict binary safe/unsafe labels directly from image observations. \textbf{Vision Language Models (VLMs):} We use Qwen3.5 9B~\citep{qwen3.5}
and DeepSeek-VL2 4B~\citep{wu2024deepseek} to classify each observation as safe or unsafe. 
\textbf{VLMs + in-context learning (ICL):} We provide Qwen3.5 9B~\citep{qwen3.5} and DeepSeek-VL2 4B~\citep{wu2024deepseek} with task-specific classification rules in the prompt and ask them to infer whether each scene violates the corresponding safety constraint. \textbf{Tensor-based Reasoner:} Furthermore, we compare \gls{neupro} with a tensor-based variant that adopts tensor reasoner~\citep{shindo2023alpha} as its backbone. Experiments are running on a RTX A6000 GPU with 48GB RAM. Architectural details of the black-box classifier and MLP within \gls{neupro} are in App.~C.
\begin{figure*}
    \includegraphics[width=\textwidth]{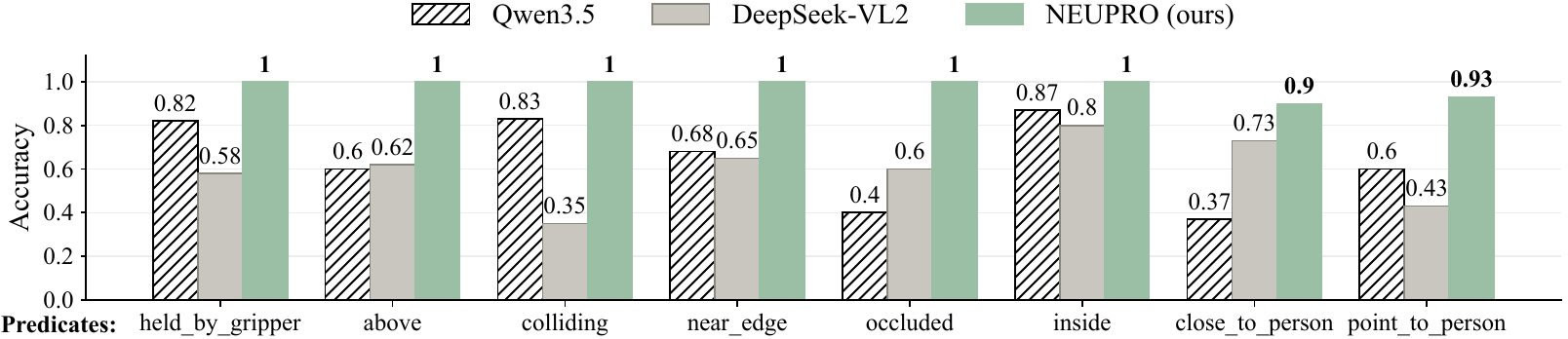}
\caption{\textbf{\gls{neupro} accurately learns predicates through differentiable reasoner supervision from raw images.} \gls{neupro} outperforms baseline methods in predicate grounding accuracy. For readability, only mean accuracy is shown. Details see RQ1.   
}
    \label{fig:rq1}
\end{figure*}
\begin{table*}[t]
\centering
\resizebox{\textwidth}{!}{%
\begin{tabular}{lcccccccc}
\toprule
\multirow{2}{*}{\textbf{\texttt{Method}}}
& \multicolumn{8}{c}{\textbf{\texttt{Tasks}}} \\
\cmidrule(lr){2-9}
& \textbf{\texttt{Occlusion}}
& \textbf{\texttt{Cabinet}}
& \textbf{\texttt{Near}}
& \textbf{\texttt{Above}}
& \textbf{\texttt{Collision}}
& \textbf{\texttt{Close}}
& \textbf{\texttt{Pointing}} 
& \textbf{\texttt{All task}} \\
\midrule
\texttt{Black-box classifier}
& 0.84\spm{0.09} & \textbf{0.96\spm{0.04}} & 0.79\spm{0.08} & 0.98\spm{0.05} & 0.96\spm{0.05}  & 0.98\spm{0.04} & 0.86\spm{0.14} & \textbf{\texttt{N/A}} \\
\texttt{DeepSeek-VL2 (ICL)}
& 0.41\spm{0.11} & 0.5\spm{0.11} & 0.54\spm{0.07} & 0.3\spm{0.24} & 0.31\spm{0.07} & 0.4\spm{0.06} & 0.33\spm{0.16} & 0.48\spm{0.03} \\

\texttt{Qwen3.5 (ICL)}
& 0.49\spm{0.02} & 0.61\spm{0.02} & 0.39\spm{0.02} & 0.68\spm{0.16} & 0.52\spm{0.14}  & 0.82\spm{0.04} & 0.57\spm{0.16} & 0.55\spm{0.04} \\

\texttt{DeepSeek-VL2}
& 0.41\spm{0.01} & 0.51\spm{0.06} & 0.53\spm{0.04} & 0.31\spm{0.09} & 0.39\spm{0.02} & 0.21\spm{0.06} & 0.31\spm{0.03} & 0.44\spm{0.04} \\

\texttt{Qwen3.5}
& 0.5\spm{0.02} & 0.62\spm{0.01} & 0.39\spm{0.0} & 0.78\spm{0.03} & 0.75\spm{0.0} & 0.8\spm{0.11} & 0.77\spm{0.02} & 0.55\spm{0.04} \\

\textbf{\texttt{NEUPRO (ours)}}
& \textbf{1.0\spm{0.0}}
& \textbf{0.96\spm{0.06}}
& \textbf{0.99\spm{0.03}}
& \textbf{1.0\spm{0.0}}
& \textbf{1.0\spm{0.0}}
& \textbf{1.0\spm{0.0}}
& \textbf{0.92\spm{0.08}} & \textbf{0.92\spm{0.02}} \\
\bottomrule
\end{tabular}
}
\caption{\textbf{\gls{neupro} accurately infer safety outcomes from raw visual observation.} Safety classification accuracy comparison across all REASON tasks with baseline methods (the higher, the better, best performing bolded). Results averaged over five test groups with std, details see RQ2. ICL denotes in-context learning.}
\label{tab:rq2}
\end{table*}
\paragraph{RQ1:} \textbf{\gls{neupro} accurately learns safety-relevant predicates from raw images through differentiable reasoner supervision.} 
We evaluate the accuracy of eight predicates learned by \gls{neupro}. For each predicate, we use five test images and compute the mean accuracy of the corresponding grounded atoms. Fig.~\ref{fig:rq1} compares \gls{neupro} with Qwen3.5 and DeepSeek-VL2 across all eight predicates. \gls{neupro} achieves the highest accuracy, whereas the VLM baselines exhibit substantially lower and less consistent performance, particularly on relational (e.g., \textbf{\texttt{occluded}}) and human-interaction predicates (e.g., \textbf{\texttt{point\_to\_human}}). These results show that safety supervision propagated through the differentiable reasoner effectively guides the perception module to ground accurate and safety-relevant predicates from visual observations. Prompts and other metrics in App.~D. 


\begin{figure}[t]
\includegraphics[width=\columnwidth]{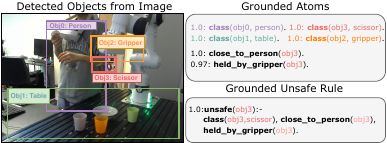}
    \caption{\textbf{Qualitative example of safety violation explanations produced by NEUPRO}, in the form of first-order logic rules and atoms. In this scene, the grounded atoms indicate that the robot is holding scissors and is close to a person, from which \gls{neupro} infers an unsafe prediction through the corresponding first-order logic rule. 
    }
    \label{fig:rq3}
\end{figure}
\paragraph{RQ2:} \textbf{\gls{neupro} accurately infers safety outcomes directly from raw visual observations.} Building on the learned safety-relevant predicates, we next evaluate whether \gls{neupro} can compose them through symbolic rules to correctly determine whether a scene satisfies or violates a safety constraint. We conduct this evaluation across all seven REASON tasks and report task accuracy, defined as the percentage of test scenes for which the predicted safety outcome is correct. Tab.~\ref{tab:rq2} compares \gls{neupro} with task-specific black-box MLP classifiers, VLM-based baselines, and VLM-based in-context learning(ICL) baselines. The MLP baseline is trained separately for each task and therefore does not support evaluation in the joint all-task setting. In contrast, \gls{neupro} can be adapted to additional tasks by extending the rule base without redesigning the underlying safety-reasoning mechanism. All results are averaged over five test groups and reported with standard deviations. Overall, \gls{neupro} achieves consistently strong accuracy across tasks, demonstrating that its learned predicates can be effectively composed for end-to-end safety reasoning from raw images. Although VLMs can be applied across multiple tasks, they achieve substantially lower accuracy (even when task rules are provided in context), highlighting the advantage of explicit predicate grounding and symbolic reasoning. We provide the prompt, the black-box classifier architecture, and other metrics in App.~E.
\begin{figure}[t]
    \includegraphics[width=\columnwidth]{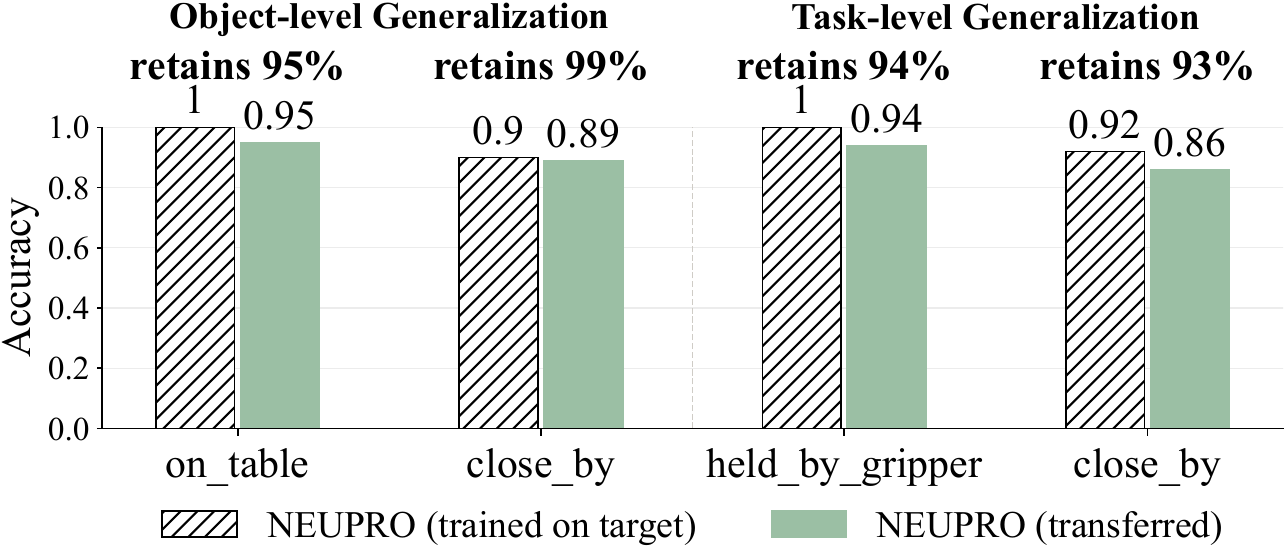}
    \caption{\textbf{\gls{neupro} learned predicates are transferrable.} Object-level generalization evaluates whether predicates remain accurate when applied to different object instances or categories. Task-level generalization compares predicates trained directly on a target task with predicates transferred from another task. \gls{neupro} retains high accuracy under both settings, demonstrating that its learned representations are reusable across objects and tasks. Details see RQ4.
    }
    \label{fig:rq4}
\end{figure}
\paragraph{RQ3:} \textbf{\gls{neupro} can explain safety violations.} A key advantage of \gls{neupro} is its ability to provide interpretable safety explanations. We now qualitatively demonstrate this capability. Unlike black-box classifiers that only output a safe/unsafe label, \gls{neupro} infers safety by composing atoms through interpretable logic rules. As shown in Fig.~\ref{fig:rq3}, the scene is predicted as unsafe because the grounded rule is satisfied by atoms indicating that the robot gripper is holding a scissor and that the scissor is close to a person.

\begin{table}[t]
\centering
\small
\setlength{\tabcolsep}{1.6pt}
\begin{tabular*}{\columnwidth}{
    @{\extracolsep{\fill}}
    l
    cccc
    @{}
}
\toprule
\multirow{2}{*}{\textbf{\texttt{Method}}}
& \multicolumn{2}{c}{\textbf{\texttt{Above\_Laptop}}}
& \multicolumn{2}{c}{\textbf{\texttt{Close\_Person}}} \\
\cmidrule(lr){2-3}
\cmidrule(lr){4-5}
& \textbf{\texttt{Hard}}
& \textbf{\texttt{Soft}}
& \textbf{\texttt{Sharp}}
& \textbf{\texttt{N.sha}} \\
\midrule
\textbf{\texttt{\gls{neupro} (ours)}}
& \textbf{1.0\spm{0.0}}
& \textbf{1.0\spm{0.0}}
& \textbf{1.0\spm{0.0}}
& \textbf{1.0\spm{0.0}} \\

\texttt{Qwen3.5}
& 0\spm{0.0}
& 0.96\spm{0.08}
& 0.04\spm{0.08}
& 0.96\spm{0.08} \\

\texttt{Qwen3.5(ICL)}
& 0.36\spm{0.15}
& 0.76\spm{0.08}
& 0.6\spm{0.28}
& 0.92\spm{0.10} \\

\texttt{DeepSeek-VL2}
& 0.64\spm{0.15}
& 0.32\spm{0.20}
& 0.72\spm{0.16}
& 0.36\spm{0.08} \\

\texttt{DeepSeek-VL2(ICL)}
& 0.56\spm{0.29}
& 0.44\spm{0.32}
& 0.84\spm{0.08}
& 0.4\spm{0.22} \\
\bottomrule
\end{tabular*}

\caption{\textbf{\gls{neupro} supports flexible safety reasoning with
commonsense knowledge.}
We evaluate whether each method distinguishes identical spatial relations
with different object properties: hard vs.\ soft objects above a laptop,
and sharp vs.\ non-sharp objects close to a person. Results are reported
as mean accuracy with std. See RQ5 for details.}
\label{tab}
\end{table}

\paragraph{RQ4:} \textbf{\gls{neupro} learned predicates can be transferred to different objects and tasks.} 
Since \gls{neupro} grounds safety concepts as semantic predicates rather than task-specific labels, predicates learned in one context can be reused by other safety rules that require the same underlying concepts. We evaluate predicate transfer under two settings. \textbf{(i) object-level generalization} measures whether a predicate learned from one set of object instances or categories remains accurate when applied to unseen objects. \textbf{(ii) task-level generalization} evaluates whether a predicate learned in one task can be reused in a different task that requires the same underlying semantic concept e.g., predicates such as 
\textbf{\texttt{closeby}} may appear in multiple tasks involving different objects or interaction scenarios. 
Fig.~\ref{fig:rq4} summarizes the accuracy results over five test groups, details and other metrics see App.~F. In both settings, \gls{neupro} is evaluated on tasks/objects that differ from those used during training. Results show that \gls{neupro} maintains strong performance under such transfer settings, suggesting that it learns reusable safety-relevant representations rather than task-specific visual shortcuts.

\paragraph{\textbf{RQ5:}} \textbf{\gls{neupro} can reason about safety specifications beyond fixed object-relation associations by incorporating commonsense knowledge.}
We evaluate this capability in two tasks in which the same spatial relation yields different safety outcomes depending on the manipulated object. In the first task, holding a hard object, e.g., a cup, above a laptop is unsafe, whereas holding a soft object, e.g., a towel, in the same position is safe. In the second task, holding a sharp object close to a person is unsafe, whereas a non-sharp object in the same spatial configuration is safe. Solving these tasks requires grounding the relevant visual relations and combining them with background knowledge about object attributes. Tab.~\ref{tab} reports accuracy results over five test groups, details, and other metrics in App. G. 
Results show that \gls{neupro} supports flexible safety reasoning and avoids relying on fixed visual associations between relations and safety labels.  

\begin{figure}[t]
    \includegraphics[width=\columnwidth]{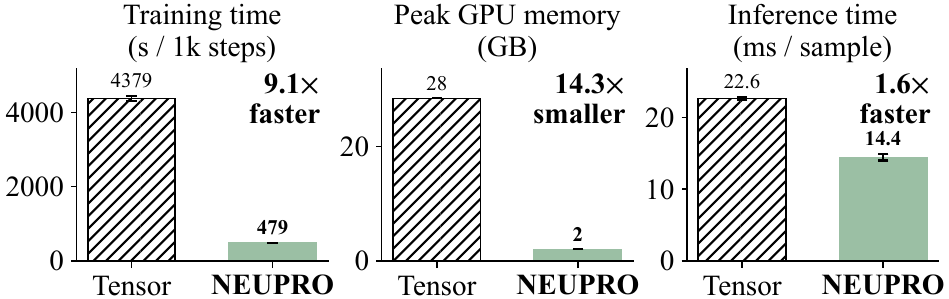}
    \caption{\textbf{\gls{neupro} supports scalable training and inference through graph-based reasoning.} We compare graph-based \gls{neupro} with its tensor-based variant. 
    \gls{neupro} achieves \textbf{$9.1\times$ faster} training, \textbf{$14.3\times$ lower} memory usage and \textbf{$1.6\times$ faster} inference, demonstrating substantially better scalability. Details see RQ6.  
    }
    \label{fig:rq6}
\end{figure}

\paragraph{\textbf{RQ6:}} \textbf{\gls{neupro} supports scalable safety predicate learning and inference due to its graph-based differentiable reasoning.} We compare graph-based \gls{neupro} with its tensor-based variant using all tasks in REASON, and evaluate both methods in terms of training time, peak GPU memory usage, and inference time. Fig.~\ref{fig:rq6} summarizes the comparison results. While achieving the same accuracy, graph-based \gls{neupro} yields \textbf{$9.1\times$ faster} training, \textbf{$14.3\times$ lower} peak GPU memory usage, and \textbf{$1.6\times$ faster} inference. Note that we use 102 rules in this evaluation. As the number of rules increases, the efficiency advantage of graph-based reasoning is expected to become even larger.

\section{Limitations}
\label{sec:limitation}
Although \gls{neupro} demonstrates strong capability for interpretable and flexible safety reasoning, several limitations remain and open promising directions for future work. First, \gls{neupro}'s reasoning performance depends on the quality and coverage of the safety rules and background knowledge. If a safety condition is missing or the knowledge base lacks the relevant object properties, \gls{neupro} may fail to identify the corresponding violation. 
Second, \gls{neupro} currently focuses on visual safety reasoning from static observations, and its performance relies on the underlying vision backbone grounding DINO. When grounding DINO fails to detect an object, \gls{neupro} lacks information about that object. 
Third, many real-world safety constraints are inherently temporal, involving motion trends or future consequences of an action. Incorporating temporal predicates and predictive world models would allow \gls{neupro} to reason about evolving safety risks, such as whether a moving object is likely to fall or collide.
Fourth, although the REASON dataset contains images from multiple viewpoints, only one image is used during inference. Incorporating multi-view reasoning could improve the model’s robustness, particularly when a single view is affected by occlusion.

\section{Conclusion}
\label{sec:conclusion}

We presented \gls{neupro}, a neuro-symbolic framework for interpretable and flexible robot safety reasoning from visual observations. \gls{neupro} alleviates the non-interpretable and inflexible limitations of previous work by representing safety specifications as interpretable first-order logic rules. \gls{neupro} supports scalable predicate learning and safety reasoning from raw images due to its graph-based reasoning. 
Furthermore, we introduced \emph{REASON}, the first real-robot benchmark for interpretable safety specification. 
Experiments on \emph{REASON} validate \gls{neupro}'s interpretability, flexibility, and scalability.  
Overall, \gls{neupro} provides a transparent and reusable alternative to task-specific scalar costs and opaque safety representations. By connecting raw perception, symbolic knowledge, and differentiable reasoning, \gls{neupro} provides a promising foundation for robot learning systems that deliver accurate, interpretable, transferable, and scalable safety decisions.

\section{Acknowledgements}
\label{ackowledgements}
This work was funded by the Deutsche Forschungsgemeinschaft (DFG, German Research Foundation) under Germany's Excellence Strategy – EXC-3066, ``The Adaptive Mind'', and EXC-3057, ``Reasonable AI''. It was also funded by the German Federal Ministry of Education and Research, the Hessian Ministry of Higher Education, Research, Science and the Arts (HMWK) within their joint support of the National Research Center for Applied Cybersecurity ATHENE, via the ``SenPai: XReLeaS'' project. This work also benefits from DFG Emmy Noether Programme (CH 2676/1-1), the EU Horizon Europe projects MANiBOT (101120823) and ARISE (101135959), the BMFTR project RIG (16ME1001), and the ERC project SIREN (101163933). We also acknowledge support from the hessian.AI Service Center (BMFTR, 16IS22091), the hessian.AI Innovation Lab (S-DIW04/0013/003), Google, and the Alfried Krupp Foundation.

\bibliography{aaai2027}


\end{document}